\documentclass{frontiersSCNSmod}
\usepackage{inputenc,hyperref, booktabs, amsfonts, nicefrac, url,nicefrac, algorithm, algorithmic,lineno,microtype,subcaption}
\usepackage[onehalfspacing]{setspace}

\def\firstAuthorLast{Travnik {et~al.}}
\def\Authors{Jaden Travnik\,$^{1,2}$, Kory W. Mathewson\,$^{1,2}$, Richard S. Sutton\,$^{1}$ and Patrick M. Pilarski\,$^{1,2,*}$}

\def\Address{$^{1}$Reinforcement Learning and Artificial Intelligence Laboratory, Department of Computing Science, University of Alberta, Edmonton, AB, Canada;\\$^{2}$Bionic Limbs for Improved Natural Control Laboratory, Department of Medicine, University of Alberta, Edmonton, AB, Canada}

\def\corrAuthor{Patrick M. Pilarski, Division of Physical Medicine and Rehabilitation, Department of Medicine, 5-005 Katz Group Centre for Pharmacy and Health Research, University of Alberta, Edmonton, Canada, T6G 2E1.}

\def\corrEmail{pilarski@ualberta.ca}

\newcommand{\INDSTATE}[1][1]{\STATE\hspace{#1\algorithmicindent}}

\begin{document}
\onecolumn
\firstpage{1}

\title[Reactive Reinforcement Learning]{Reactive Reinforcement Learning in Asynchronous Environments}

\author[\firstAuthorLast ]{\Authors}
\address{}
\correspondance{}

\extraAuth{}

\maketitle

\begin{abstract}
The relationship between a reinforcement learning (RL) agent and an asynchronous environment is often ignored. Frequently used models of the interaction between an agent and its environment, such as Markov Decision Processes (MDP) or Semi-Markov Decision Processes (SMDP), do not capture the fact that, in an asynchronous environment, the state of the environment may change during computation performed by the agent. In an asynchronous environment, minimizing reaction time---the time it takes for an agent to react to an observation---also minimizes the time in which the state of the environment may change following observation. In many environments, the reaction time of an agent directly impacts task performance by permitting the environment to transition into either an undesirable terminal state or a state where performing the chosen action is inappropriate. We propose a class of {\em reactive reinforcement learning algorithms} that address this problem of asynchronous environments by immediately acting after observing new state information. We compare a reactive SARSA learning algorithm with the conventional SARSA learning algorithm on two asynchronous robotic tasks (emergency stopping and impact prevention), and show that the reactive RL algorithm reduces the reaction time of the agent by approximately the duration of the algorithm's learning update. This new class of reactive algorithms may facilitate safer control and faster decision making without any change to standard learning guarantees.

\end{abstract}

\section{Introduction}

Reinforcement learning (RL) algorithms for solving optimal control problems are comprised of four distinct components: acting, observing, choosing an action, and learning. This ordering of components forms a protocol which is used in a variety of applications. Many of these applications can be described as synchronous environments where the state of the environment remains in the same state until the agent acts at which point the environment immediately returns its new state. In these synchronous environments, such as Backgammon [Tesauro 1994] or classic control problems, it is not necessary to know the computation time to perform any of the protocol's components. For this reason, most reinforcement learning software libraries, such as RL-Glue [Tanner and White 2009], BURLAP \footnote{http://burlap.cs.brown.edu/} or OpenAI gym \footnote{https://gym.openai.com/}, have functions which accept the agent's action, and return the new state and reward immediately. These functions remain convenient for simulated environments where the dynamics of the environment can be computed easily [Sutton and Barto 1998]. However, unlike synchronous environments, asynchronous environments do not wait for an agent to select an action before they change state. The computation of RL protocol components (acting, observing, choosing an action, learning) takes time and an asynchronous environment will continually change state during this time [Degris and Modayil 2012, Hester et al. 2012, Caarls and Schuitema 2015]. This can negatively affect the performance of the agent. If the agent's reaction time is too long, its chosen action may become inappropriate in the now changed environment. Alternatively, the environment may have moved into an undesirable terminal state.

In this paper, we explore a very simple alternative arrangement of the reinforcement learning protocol components. We first investigate a way to reorder SARSA control algorithms so that they are able to react to the most recent observation before learning about the previous time step; we then discuss convergence guarantees of these reordered approaches when viewed in discrete time (following Singh et al.\ [2000]). Then, we examine a asynchronous continuous-time robot task where the reaction times of agents affect the overall task performance---in this case, breaking or not breaking an egg with a fast-moving robotic arm. Finally, we present a discussion on the implementation of reactive algorithms and their application in related settings. 

\subsection{Related Background}

The focus of most contemporary RL research is on action selection, representation of state, and the learning update itself; the performance impact of reaction time is considered less frequently, but is no less important of a concern [Barto et al. 1995]. Several groups have discussed the importance of minimizing reaction time [Degris and Modayil 2012, Hester et al. 2012, Caarls and Schuitema 2015]. Hester et al. noted that existing model-based reinforcement learning methods may take too much time between successive actions and presented a parallel architecture that outperformed traditional methods. Caarls and Schuitema extended this parallel architecture to the online learning of a system's dynamics [Caarls and Schuitema, 2015]. Their learned model allowed for the generation of simulated experience which could be combined with real experience in batch updates. While parallelization methods may improve performance, they are computationally demanding. We propose an alternative approach when system resources are constrained.

\section{Temporal delays in Asynchronous Environments}

Temporal-difference (TD) control algorithms like SARSA and Q-Learning [Sutton and Barto 1998, Watkins and Dayan 1992] were introduced with synchronous discrete-time environments in mind; these environments are characterized by remaining stationary during the planning and learning of the agent. In synchronous environments, the time to perform the individual components of the SARSA algorithm protocol has no impact on task performance. Specifically, the time it takes to react to a new game state in chess has no influence over the end of the game. In asynchronous environments however, the time it takes for the agent to react to new observations can drastically influence its performance on the task. Such as, in the original formulation of a cart-pole, the agent applied it's actions {em\ left and right} at discrete time intervals [Barto et al. 1983]. These time intervals were set small enough so that the pole would not fall further than the agent would be able recover.

\begin{figure}[!ht]
\begin{center}
\includegraphics[angle=270,origin=c,width=.2\textwidth]{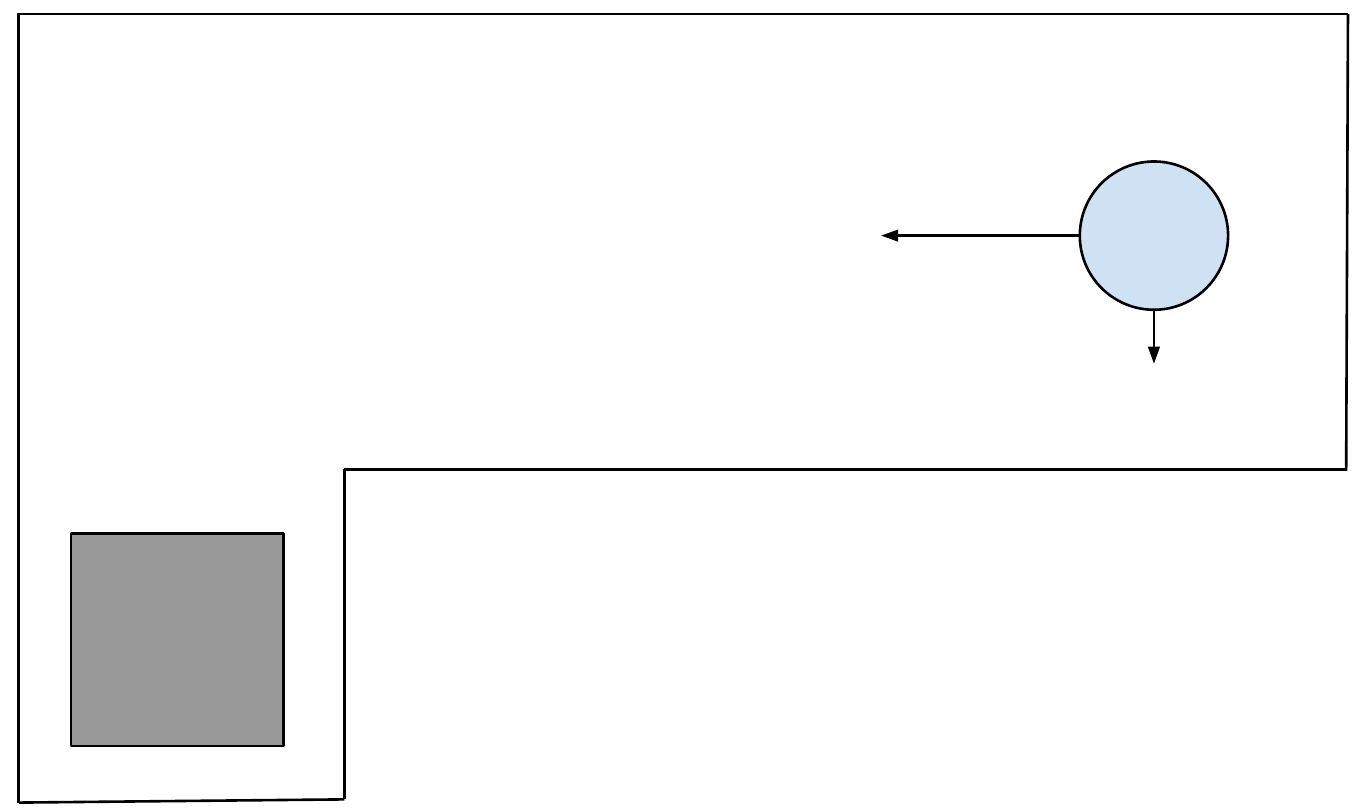}
\caption{The Hallway World task with the agent (the blue circle) starting near the bottom of the hallway. The gray square denotes the terminal state. The arrows denote the 2 actions which move the agent leftwards or upwards and continue moving the agent in that direction until interrupted.}

\label{fig:hallway}
\end{center}
\end{figure}

As a concrete example, imagine an asynchronous environment we here call Hallway World with a left turn leading to the terminal state, as shown in Fig. \ref{fig:hallway}. The agent starts an episode near the bottom of a hallway and has two actions: \textit{move left} and \textit{move up} which move the agent in a direction and continue to move the agent in that direction with constant velocity until interrupted with the other action, hitting a wall, or arriving in the terminal state. If the agent hits a wall it receives a reward of -1 and comes to a stop. When the agent reaches the terminal state it will receive a negative reward directly proportional to the duration of the episode. In this way, the agent is motivated to get to the terminal state as quickly as possible without touching the walls. The only observation that the agent can make is to determine if there is a wall on its left. 

The optimal policy in Hallway World is for the agent to \textit{move upward and observe the wall continually until an opening in the wall is observed then immediately move leftwards towards the terminal state}. SARSA (Alg. \ref{algo:sarsa}) is unable to learn this optimal policy because it is restricted by the delay between observing the opening in the wall and moving towards the terminal state. Even in the best case scenario, that is, if the SARSA agent observed the opening in the wall just as it appeared, it would not be able to act on this observation until it had spent time learning about the previous action and observation. Assuming that the components of the algorithm (acting, observing, choosing an action, and learning) each take some constant amount of time $t_c$, if a SARSA agent observes an opening in the wall, it must choose to move left and learn about the previous state-actions before taking the action, this would add $2t_c$ onto episode time, thereby affecting the total reward and task performance. Thus, overall performance in Hallway World decreases with the time the agent spends selecting an action and learning, irrespective of how these components are performed.

\section{Reactive SARSA}

To minimize the time between observing a state and acting upon it, we propose a modification to conventional TD-control algorithms: \textit{take actions immediately after choosing them given the most recent observation}. We propose a straightforward new algorithm, Reactive SARSA, as one example of this modification (Alg. \ref{algo:switched-sarsa}); in each step of the learning loop, the agent observes a reward and new state, chooses an action from a policy based on the new state, immediately takes that action, then performs the learning update based on the previous action.

\begin{algorithm}
\begin{algorithmic}
\STATE Initialize $Q(s,a)$ arbitrarily, for all $s \in \mathcal{S}, a \in \mathcal{A}(s)$
\STATE Repeat (for each episode):
  \INDSTATE[1] Initialize $S$
    \INDSTATE[1] Choose $A$ from $S$ using policy derived from $Q$ (e.g., $\epsilon$-greedy)
    \INDSTATE[1] Repeat (for each step of episode):
    \INDSTATE[2] Take action $A$
        \INDSTATE[2] Observe $R$, $S'$
        \INDSTATE[2] Choose $A'$ from $S'$ using policy derived from $Q$ (e.g., $\epsilon$-greedy)
        \INDSTATE[2] $Q(S,A)$ $\leftarrow$ $Q(S,A)$ + $\alpha$[$R$ + $\gamma$$Q(S',A')$ - $Q(S,A)$]
        \INDSTATE[2] $S$ $\leftarrow$ $S'$, $A$ $\leftarrow$ $A'$
\end{algorithmic}
\caption{SARSA: An on-policy TD control algorithm}
\label{algo:sarsa}
\end{algorithm}

\begin{algorithm}
\begin{algorithmic}
\STATE Initialize $Q(s,a)$ arbitrarily, for all $s \in \mathcal{S}, a \in \mathcal{A}(s)$
\STATE Repeat (for each episode):
  \INDSTATE[1] Initialize $S$
    \INDSTATE[1] Choose $A$ from $S$ using policy derived from $Q$ (e.g., $\epsilon$-greedy)
    \INDSTATE[1] Take action $A$
    \INDSTATE[1] Repeat (for each step of episode):
        \INDSTATE[2] Observe $R$, $S'$
        \INDSTATE[2] Choose $A'$ from $S'$ using policy derived from $Q$ (e.g., $\epsilon$-greedy)
    \INDSTATE[2] Take action $A'$        
        \INDSTATE[2] $Q(S,A)$ $\leftarrow$ $Q(S,A)$ + $\alpha$[$R$ + $\gamma$$Q(S',A')$ - $Q(S,A)$]
        \INDSTATE[2] $S$ $\leftarrow$ $S'$, $A$ $\leftarrow$ $A'$
\end{algorithmic}
\caption{Reactive SARSA: A \textit{reactionary} on-policy TD algorithm}
\label{algo:switched-sarsa}
\end{algorithm}

The slight reordering of RL algorithm protocol components does not effect convergence in discrete time. Here, we provide a basic theoretical sketch that, in discrete-time synchronous tasks, Reactive SARSA learns the same optimal policy as SARSA, in the same manner. As is illustrated in Fig. 2, this equivalence is trivially evident by observing that in both algorithms the first 2 actions are selected using the initial policy. In each subsequent step $t$, actions are chosen using the policy learned on the last step, and the policy updates happen with identical experiences (Fig. 2).

\begin{figure}[!th]
\begin{center}
\includegraphics[width=.9\textwidth]{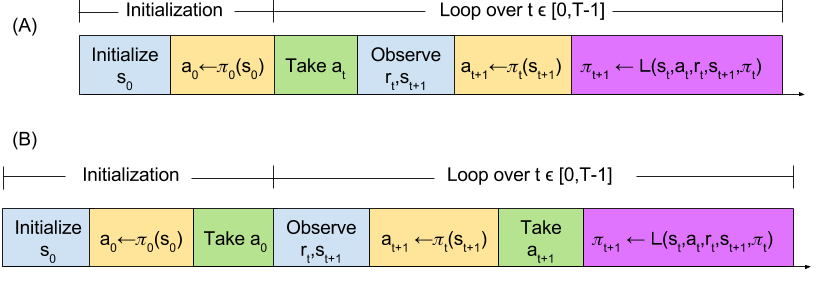}
\caption{Time step comparison of (A) standard and (B) reactive reinforcement learning algorithms. The function L refers to a learning function which updates the policy $\pi$. The learning function L is not limited to the SARSA learning update and encompasses any learning update such as Q-learning.}
\label{fig:alg-comparison}
\end{center}
\end{figure}

If we redefine Hallway World as a synchronous environment where the agent moves a constant distance for each action instead of continually moving, the same policies and performance would be expected between both algorithms and this is what we found in practice. The difference between reactive and non-reactive algorithms is the order of the RL components (acting, observing, choosing an action, and learning).

\section{Experiments}

To explore the differences between the SARSA and Reactive SARSA learning algorithms in asynchronous environments, we designed a reaction-time-dependent task with similar qualities to the Hallway World described above and illustrated in Fig.\ \ref{fig:hallway}. The task was performed using one joint of a robotic arm (an open-source robotic arm [Dawson et al. 2014] shown in Fig. \ref{fig:robot}). We conducted two experiments with the same episodic stopping task. The arm started at one extreme of the joint rotation range and was then rotated quickly towards the other end of its range. The agent needed to stop the rotation as soon as possible following an indication to stop that was observed by a state change from ``Normal'' to ``Emergency''.

The agent had two actions: \textit{stop} and \textit{move}. If the agent chose to \textit{stop} while in the ``Normal'' state, the agent would receive a constant reward of -1, remain in the ``Normal'' state, and the arm would continue rotating. If the agent chose to \textit{move} while in the ``Normal'' state, the agent would receive a reward of 0, remain in the ``Normal'' state, and continue rotating. Once the ``Emergency'' state had been observed, the reward for either action would be a negative reward proportional to the amount of time (in $\mu s$) spent in the ``Emergency'' state. When the agent chose to \textit{stop} in the ``Emergency'' state, it transitioned to the terminal state, thereby ending the episode. This reward definition was chosen as a convenient means of valuing reaction time; the distance traveled during the reaction time would also have been a valid alternative. 

If complete information about the stopping task was available, optimal performance could be obtained through direct engineering of a control system designed to stop the arm as soon as the state changes. However, the agent did not know which state was the ``Emergency'' state and used its experience to learn what to do in any given state. By excluding the complex state information of many real-world robotic tasks and using this simple stopping task, we were able to investigate the effects of reaction time on overall performance and the differences between conventional and reactive TD-control algorithms.

\subsection{Experiment 1}

\begin{figure}[!t]
\begin{center}
\includegraphics[height=2.2in]{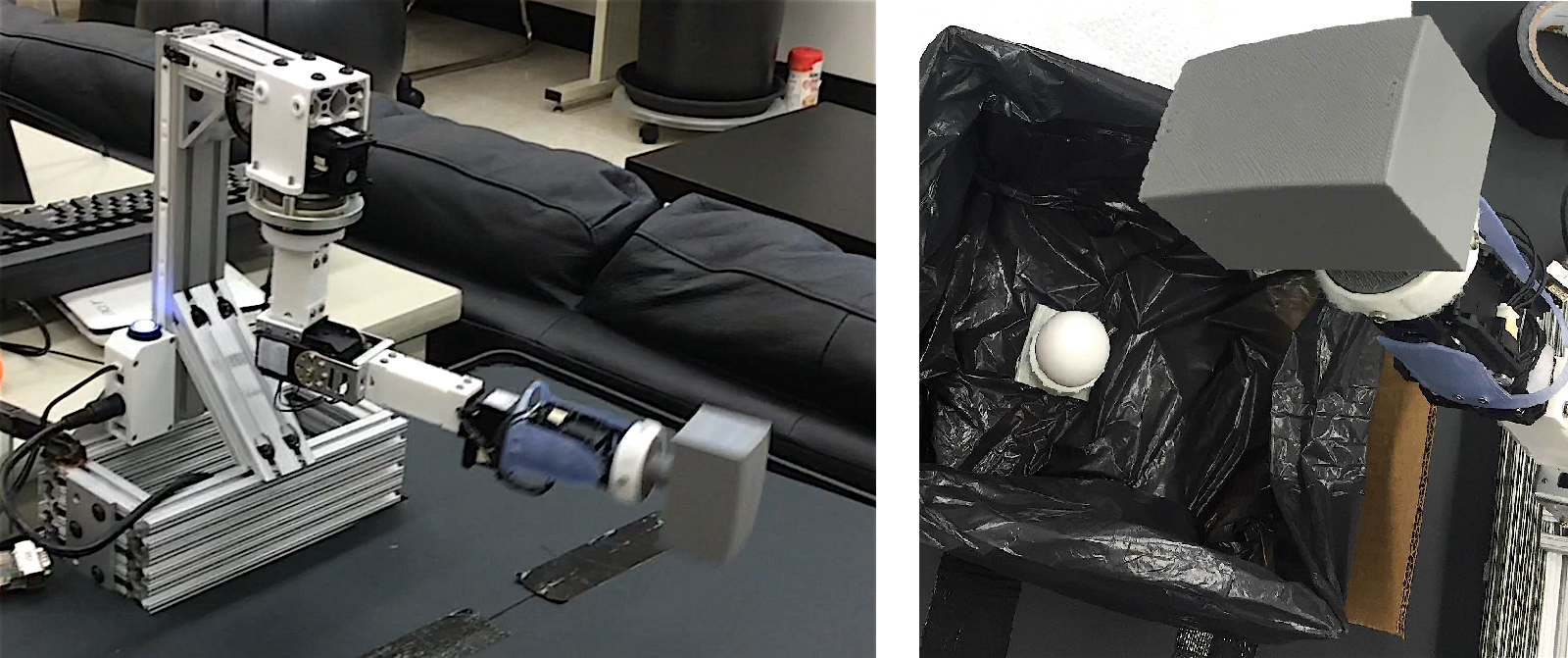}
\caption{Experimental setup, showing the robot arm in motion for the first experiment (left) and the robot arm poised to impact an egg during the second experiment (right).}
\label{fig:robot}
\end{center}
\end{figure}

To explore the effect of the reactive algorithms on reaction time and task performance, the robotic arm was programmed to move at a constant velocity along a simple trajectory (Fig. \ref{fig:robot}, left). The experiment involved 30 trials, each of which was comprised of 20 episodes with the agent starting in a ``Normal'' state and switching to the ``Emergency'' state after some random amount of time. The standard and Reactive SARSA agents were compared with greedy policies, $\gamma$ = 0.9, $\lambda$ = 0.9, and $\alpha$ = 0.1.

 There are many potential causes for time delays in the learning step. One example comes from the idea of predictive knowledge representations. Here knowledge is represented and learned as a collection of predictions about a robots observed experience. Such knowledge may be updated and computed during each cycle. One approach to building this knowledge is the Horde architecture, Horde introduces the idea of \textit{demons} which learn predictions about the environment and can build on each other to achieve a scalable method of knowledge learning [Sutton et al. 2011].
 A Horde architecture with 2576 demons (predicting the position, velocity, temperature, load and other measures) was experimentally validated on the robotic arm. On the experimental setting tested this setup resulted in an average computation time of one demon's prediction to be ~3.33 $\mu$s. The more predictions one wants to make, the longer the duration of learning, thus the reaction time increases. Specifically, the time delays of 50ms, 100ms, 250ms, and 500ms on the experimental hardware are equivalent to a horde architecture of approximately 15000, 45000, 75000, and 150000 demons, respectively. It is clear that more predictions increase the reaction-time, and the addition of time delays in the following experiments was used to appropriately simulate the addition of more predictions. To simulate the performance of these additional predictions and modulate in a controlled fashion the effect of longer learning steps, these time delays were added to the learning update step.

\begin{figure}[!t]
\begin{center}
\includegraphics[width=.75\textwidth]{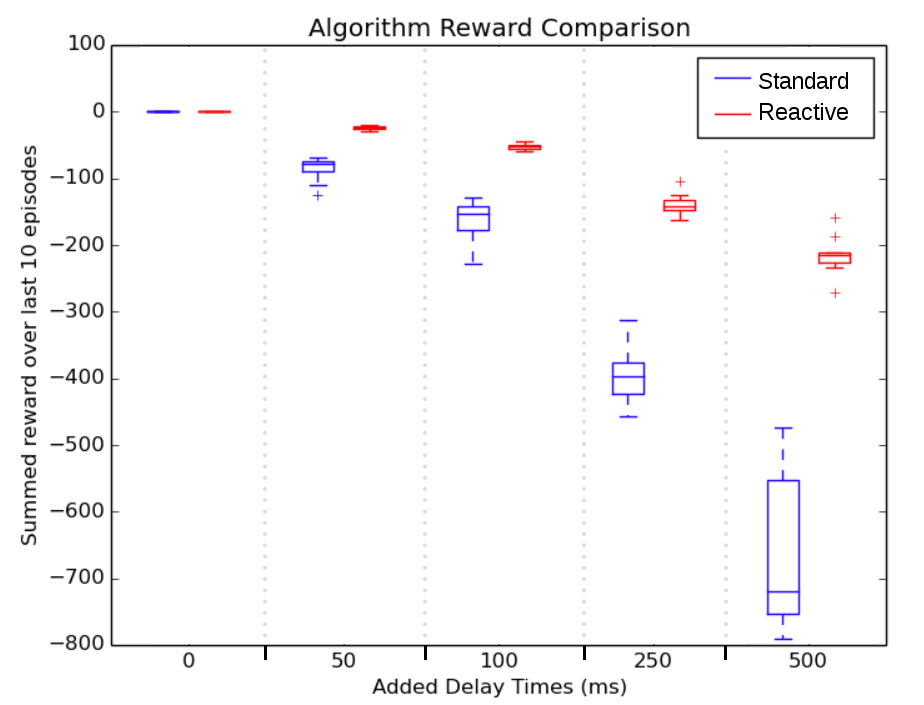}
\caption{Key result: a comparison of summed reward over the last 10 episodes of 30 trials across 5 different learning delay length lengths during robot arm motion. Reactive SARSA had a significantly reduced reaction time when compared to the standard SARSA algorithm for all delay lengths. }
\label{fig:reward-comparison}
\end{center}
\end{figure}

Figure \ref{fig:reward-comparison} shows how the the duration of learning influenced the task performance. The figure shows the average cumulative episodic return for the last 10 episodes, once both agents had learned policies. As the delay increased, both algorithms suffered performance decreases, but the standard SARSA algorithm performed worse with larger variability. While Reactive SARSA was affected by increasing time delays, the impact was less severe. Specifically, median reaction time of the Reactive SARSA was approximately half of the added learning delay. This effect is most likely because the transition from ``Normal'' to ``Emergency'' state occurred at a uniformly random selected time and since the majority of the duration of a step comprised of learning, the state change occurred on average halfway through the learning step. This also accounts for the increasing reaction time in standard SARSA, as it must wait a full additional time step before reacting to the ``Emergency'' state.   

\subsection{Experiment 2}

The second experiment considered a human-robot interaction task which demanded cooperation between a human and robotic arm to not crush an egg (Fig.\ \ref{fig:robot}, right). The robot arm was positioned above a target, in this task an egg, and would move at a constant velocity toward the target. The human was told to press a button to stop the arm before crushing the egg, and to try to stop it as close to the egg as possible without touching it. The learning task for the RL agent was to learn to stop as soon as the participant pressed a button. For the first 10 episodes of a trial, the participant trained using a hard-wired stopping algorithm which automatically stopped the arm when the participant pressed a button. For the remaining 40 episodes of the trial, previously learned SARSA and Reactive SARSA agents were used, each algorithm was used for 20 episodes, and the algorithm used was randomly alternated on each episode. The state changed from ``Normal'' to ``Emergency'' when the participant pressed a button. All three algorithmic conditions: 1) control, 2) SARSA, and 3) Reactive SARSA, included a constant 50ms delay to simulate a longer learning step (e.g., the time it would take to update the predictions for 15000 demons). The algorithm used on a trial was hidden from the participant. Four individuals participated in the experiment, providing a total of 80 episodes of each algorithm. All participants provided informed consent as per the University's Ethics Review Board and could voluntarily end the experiment at any time if they wished.  

\begin{figure}[!t]
\begin{center}
\includegraphics[width=.75\textwidth]{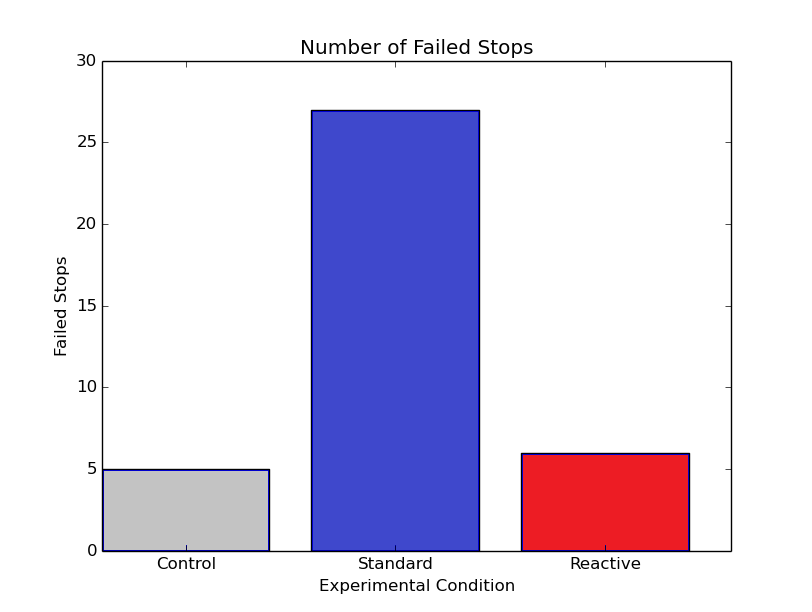}
\caption{The total number of failed stops for each algorithm during the robot's acceleration toward a breakable object (Experiment 2), summed over all four participants. For all subjects, Reactive SARSA had far fewer failed stops than standard SARSA.}
\label{fig:hit-comparison}
\end{center}
\end{figure}

Figure \ref{fig:hit-comparison} shows the total of all failed stops (``broken eggs'') for each algorithmic condition as summed across all four participants. Reactive SARSA had approximately the same number of failed stops as the optimal control strategy whereas the standard SARSA performed significantly worse with more than four times as many failed stops.

\begin{figure}[!t]
\begin{center}
\includegraphics[width=.75\textwidth]{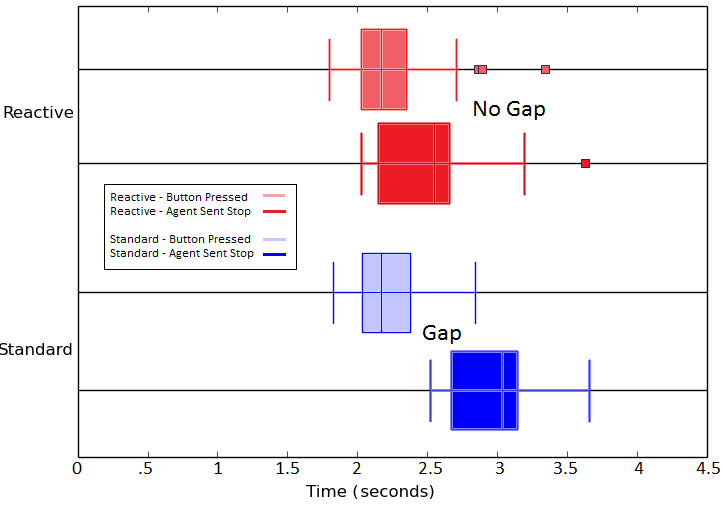}
\caption{Boxplot comparison of the distributions of events over all episodes between Reactive SARSA and the standard SARSA algorithm. Zero on the x-axis is the the moment the arm begins moving. The overlap of the button press and reactive agent's action indicates that the reactive agent has negligible delay in its reaction to the participant's input (seen in the overlap between light red and dark red, top). The standard agent's ability to act is delayed by the length of learning (visible gap between light blue and dark blue, bottom).}
\label{fig:exp2-reaction-comparison}
\end{center}
\end{figure}

In addition to comparing the number of failed stops, and thus crushed eggs, of each algorithmic condition, the time between the state change from the button push of the participant reaction time of the agent was recorded and is presented in Figure \ref{fig:exp2-reaction-comparison}. The effects of longer learning on reaction time is evident in this figure as the standard SARSA algorithm agent's Stop action is trailing behind the button press by approximately the length of the learning update and delay; this is contrasted by the tight overlap of the stimulus and action for the Reactive SARSA agent.

\section{Discussion}

Our results indicate that rearranging the fundamental components of existing TD-control algorithms (act, observe, choose action, learn) has a beneficial effect on performance in asynchronous environments where task performance is reaction-time dependent. A reactive agent can perform better in these environments as it can act immediately following observations. As would be expected, this effect becomes especially prominent as the duration of learning operations increases. Although the current experimental design added a simulated delay to the learning update step, our results indicate that as the time between observing and acting grows, performance in these environments deteriorates, regardless of the source of these delays. As standard RL algorithms perform learning and state representation construction (e.g. tile-coding [Sutton and Barto 1998], deep neural networks [Silver et al. 2016], etc.) between observing and acting, additional computation time is necessary. In asynchronous environments, as these steps become longer, the order of algorithmic components (acting, observing, choosing an action, and learning) becomes more critical. As we have shown, performance in asynchronous environments is inversely proportional to the total length of time between observations and acting.  

One alternate means of addressing delay-induced performance concerns may be to create a dedicated thread for each of the RL algorithm components (c.f., Caarls and Schuitema [2015]). We believe this is a promising area for continued research. However, as multi-threading is difficult on single-core machines such as micro-controllers, reactive algorithms as suggested in this work may have great utility when applied to embedded learning systems or smaller single-thread computers. While the order of algorithmic components might seem at first like a minor implementation detail, it may prove critical when applied to these systems. Reactive SARSA, and similar reactive algorithms, do not require multiple threads. Even in the case of multi-threaded systems, it may still prove fruitful to re-order the fundamental components of RL learning for improved performance. 

Put more strongly, we believe that allowing an RL agent to learn an optimal ordering of its learning protocol or to interrupt learning components for more pressing computations are interesting subjects of future work. As a thought experiment, imagine an oracle-agent that has perfectly modeled its environment, knowing the outcome of every possible action. If this environment is asynchronous and provides more positive rewards for completing a task as quickly as possible, then, in order for this oracle-agent to maximize its reward, it should eliminate all computations which are not necessary as they delay the agent. Since it has perfectly modeled its environment, learning does not and will not improve its model. Moreover, if by predicting the state using its perfect model, the agent can achieve a perfect state prediction without observing, observation is also an unnecessary computation. Thus the oracle-agent can eliminate learning and observing and can simply act. Experts, such as video-game speed runners or musicians, are sometimes able to perform their talents without actually observing the consequences of their actions. This is because they know their environment and task so well that they can simply act. By viewing the order of algorithmic components of learning algorithms as modifiable, an agent may be able to find an optimal ordering of its learning protocol or to interrupt long lasting computations (e.g., analyzing an image) for more pressing computations (e.g., avoiding a pedestrian).

\section{Conclusions}

RL algorithms are built on four main components: acting, observing, choosing an action, and learning. The execution of any of these components takes time, and while this may not affect synchronous discrete-time environments, it is a critical consideration for asynchronous environments, especially when task performance is proportional to the reaction time of the agent. {\em An agent should never have to wait to take an action after receiving up-to-date observations}. In this paper we present a novel reordering of the conventional RL algorithm which allows for faster reaction times. We present a simple sketch for algorithmic equivalence in synchronous discrete-time settings and show improved performance in an asynchronous continuous-time stopping task which is directly linked to agent reaction time. These results indicate that 1) reaction time is an important consideration in asynchronous environments, 2) the choice of when in a loop the RL agent should act affects an agent's reaction time, 3) reordering of the components of the algorithm as suggested here will not affect an agent's performance in synchronous discrete-time environments, 4) reactive algorithms reduce the reaction time, and thus improve performance, potentially also decreasing the time it takes for an agent to learn an optimal policy. This work, therefore, has wide potential application in real-world settings where decision making systems must swiftly respond to new stimuli.

\section*{Conflict of Interest Statement}

The authors declare that the research was conducted in the absence of any commercial or financial relationships that could be construed as a potential conflict of interest.

\section*{Author Contributions}

All authors contributed to the conception and design of the work, the interpretation of data and studies, drafting and revising the work for important intellectual content, the approval of the version to be submitted/published, and agree to be accountable for all aspects of the work.

\section*{Funding}
This research was undertaken, in part, thanks to funding from the Canada Research Chairs program, the Canada Foundation for Innovation, the Alberta  Machine Intelligence Institute, Alberta Innovates -- Technology Futures, DeepMind, and the Natural Sciences and Engineering Research Council. 

\section*{Acknowledgments}
The authors thank the other members of the Bionic Limbs for Natural Improved Control Laboratory and the Reinforcement Learning and Artificial Intelligence Laboratory for many helpful thoughts and comments.

\section*{References}

\medskip

Barto, A. G., Bradtke, S. J., and Singh, S. P. (1995). Learning to act using real-time dynamic programming. \textit{Artificial Intelligence} 72(1), pages 81-138.

Caarls, W., and Schuitema, E. (2015). Parallel On-Line Temporal Difference Learning for Motor Control. \textit{IEEE Trans. Neural Netw. Learn. Syst.}, in press.

Dawson, M. R., Sherstan, C. , Carey, J. P., Hebert, J. S., Pilarski, P. M. (2014). Development of the Bento Arm: An improved robotic arm for myoelectric training and research. In \textit{Proceedings of Myoelectric Controls Symposium (MEC)}, pages 60-64.

Degris, T., and J. Modayil. (2012). Scaling-up Knowledge for a Cognizant Robot. In \textit{Notes AAAI Spring Symposium Series.}

Hester, T., M. Quinlan, and Stone, P. (2012). RTMBA: A Real-Time Model-Based Reinforcement Learning Architecture for Robot Control. In \textit{Proceedings of the IEEE International Conference on Robotics and Automation (ICRA)}.

Kober, J., Bagnell, J. A., and Peters, J. (2013). Reinforcement Learning in Robotics: A Survey. \textit{International Journal of Robotics Research}.

Rummery, Gavin A., and Niranjan, M. (1994). On-line Q-learning using connectionist systems. Cambridge University, Cambridge, England.

Pilarski P. M., Sutton, R. S., and Mathewson, K, W. (2015) . Prosthetic Devices as Goal-Seeking Agents. In \textit{Second Workshop on Present and Future of Non-Invasive Peripheral-Nervous-System Machine Interfaces: Progress in Restoring the Human Functions}.

Singh, S., Jaakkola, T., Littman, M. L., and Szepesv\'{a}ri, C. (2000). Convergence results for single-step on-policy reinforcement-learning algorithms. \textit{Machine Learning} 38(3), pages 287-308.

Sutton, R. S., and Barto, A. G. (1998). \textit{Reinforcement learning: An introduction}. MIT press, Cambridge.\label{rl-book}

Sutton, R. S., Modayil, J., Delp, M., Degris, T., Pilarski, P. M., White, A., and Precup, D. (2011). Horde: A scalable real-time architecture for learning knowledge from unsupervised sensorimotor interaction.  In \textit{Proceedings of the 10th International Conference on Autonomous Agents and Multiagent Systems (AAMAS)}, pages 761–768.

Tanner, B., and White, A. (2009). RL-Glue: Language-Independent Software for Reinforcement-Learning Experiments. \textit{Journal of Machine Learning Research, 10(Sep), pages 2133--2136}.\label{rl-glue}

Tesauro, G. (1994). TD-Gammon, a self-teaching backgammon program, achieves master-level play. \textit{Neural computation} 6(2), pages 215-219.

Watkins, C. J. C. H., and Dayan, P. (1992). Q-learning. \textit{Machine learning} 8(3-4), pages 279-292.

Barto, A. G., Sutton, R. S., and Anderson, C. W. (1983). Neuronlike adaptive elements that can solve difficult learning control problems. \textit{IEEE transactions on systems, man, and cybernetics} 5, pages 834-846.

\end{document}